\documentclass[12pt]{article}
\usepackage{verbatim,color,amssymb,epsfig}
\usepackage[usenames,dvipsnames]{xcolor}
\usepackage{fancyhdr}
\usepackage{subcaption}
\usepackage[round, sort]{natbib}
\usepackage{algorithm}
\usepackage[normalem]{ulem}
\usepackage{amsmath}
\usepackage{algpseudocode}
\usepackage{float}
\usepackage[colorlinks, linkcolor=red, citecolor=blue]{hyperref}
\usepackage{mathtools}
\usepackage{braket}
\usepackage{geometry}
\newtheorem{Th}{\bf Theorem}

\def\boxit#1{\vbox{\hrule\hbox{\vrule\kern6pt
          \vbox{\kern6pt#1\kern6pt}\kern6pt\vrule}\hrule}}

\def\log{\hbox{log}}

\def\bse{\begin{eqnarray*}}
\def\ese{\end{eqnarray*}}
\def\be{\begin{eqnarray}}
\def\ee{\end{eqnarray}}
\def\bq{\begin{equation}}
\def\eq{\end{equation}}
\def\bse{\begin{eqnarray*}}
\def\ese{\end{eqnarray*}}

\def\b1e{{\mathbf e}}

\floatname{algorithm}{Algorithm}

\begin{document}
    \title{\LARGE Reproduction of scan B-statistic for kernel change-point detection algorithm}
\author{ Zihan Wang\\
 \small Department of Statistics and Data Science, Tsinghua University, Beijing 100084\\
{\small wangzh21@mails.tsinghua.edu.cn}
}
\date{}
\maketitle
\begin{center}
    $\mathbf{Abstract}$
\end{center}

Change-point detection has garnered significant attention due to its broad range of applications, including epidemic disease outbreaks, social network evolution, image analysis, and wireless communications. In an online setting, where new data samples arrive sequentially, it is crucial to continuously test whether these samples originate from a different distribution. Ideally, the detection algorithm should be distribution-free to ensure robustness in real-world applications. In this paper, we reproduce\footnote{\url{https://github.com/Wang-ZH-Stat/SBSK.git}} a recently proposed online change-point detection algorithm based on an efficient kernel-based scan B-statistic, and compare its performance with two commonly used parametric statistics. Our numerical experiments demonstrate that the scan B-statistic consistently delivers superior performance. In more challenging scenarios, parametric methods may fail to detect changes, whereas the scan B-statistic successfully identifies them in a timely manner. Additionally, the use of subsampling techniques offers a modest improvement to the original algorithm.

\vspace{1em}

\newpage

\section{Introduction}

Sequential analysis is a statistical method that involves analyzing data as it is collected, rather than relying on a predetermined sample size. Unlike traditional statistical methods, where the sample size is fixed in advance, sequential analysis allows for ongoing data collection and analysis simultaneously. Sampling continues until a predefined stopping rule is met, typically when significant results are observed. This approach has gained widespread popularity across various aspects of statistical research, with numerous studies highlighting its effectiveness. Sequential change detection has gained a lot of attention because of its widely applications, ranging from social networks \citep{2014Detecting}, epidemic detection \citep{baron2004early} and genomic signal processing \citep{2012Change}. 

In this paper, we reproduce an influential online change-point detection algorithm proposed by \citet{li2019scan}. They introduced an efficient kernel-based statistic for change-point detection, inspired by the recently developed B-statistic. Since this detection algorithm employs nonparametric techniques, it is free from distributional assumptions, making it robust when applied to real-world data. For simplicity, we refer to this algorithm as the \emph{Scan B-Statistic with Kernel} (SBSK).

The rest of paper is organized as follows. Section~\ref{sec.method} introduces the main idea of scan B-statistic for kernel change-point detection proposed by \cite{li2019scan} and its asymptotic properties. Section~\ref{sec.exper} provides the experimental scenario for reproductions. In Section~\ref{sec.result}, we present the simulation results and discuss the effectiveness of the algorithm. Finally, we summarize this paper in Section \ref{sec.con}.

\section{Methodology}
\label{sec.method}
Assume there are two sets $X$ and $Y$, each with $n$ samples taking value on a general domain $X$, where
$X=\{x_1,x_2,...,x_n\}$ are i.i.d. with a distribution $P$, and $Y=\{y_1, y_2,...,y_n\}$ are i.i.d. with a distribution $Q$. The maximum mean discrepancy (MMD) is defined as \cite{gretton2012kernel}
$$
{\rm MMD}[\mathcal{F},P,Q]\coloneqq\sup\limits_{f\in\mathcal{F}}\left\{\mathbb{E}_{X\sim P}[f(X)]-\mathbb{E}_{Y\sim Q}[f(X)]\right\},
$$
and an unbiased estimator of ${\rm MMD}^2$ can be obtained using U-statistic
$$
{\rm MMD}_u^2[\mathcal{F},X,Y]=\frac{1}{n(n-1)}\sum_{i\neq j}^{n}h(x_i,x_j,y_i,y_j),
$$
where $h(\cdot)$ is the kernel for U-statistic and it can be defined using an reproducing kernel Hilbert space (RKHS) kernel as
$$
h(x_i,x_j,y_i,y_j)=k(x_i,x_j)+k(y_i,y_j)-k(x_i,y_j)-k(x_j,y_i).
$$
Commonly used RKHS kernel functions include the Gaussian radial basis function (RBF) $k(x,x')=\exp(-\Vert x-x'\Vert^2/2\sigma^2)$, the Laplacian RBF $k(x,x')=\exp(-\Vert x-x'\Vert/\sigma)$, where $\sigma>0$ is the kernel bandwidth, and polynomial kernel $k(x,x')=(\Braket{x,x'}+a)^d$, where $a>0$ and $d\in\mathbb{N}$ \citep{scholkopf2002learning}. Intuitively, the empirical test statistic ${\rm MMD}^2_u$ is expected to be small (close to zero) if $P = Q$, and large if $P$ and $Q$ are “far” apart. To improve computational efficiency, an alternative approach to eatimate ${\rm MMD}^2$, called the B-test, is presented by \citet{zaremba2013b}. The key idea is to partition the n samples from $P$ and $Q$ into $N$ non-overlapping blocks of size $B$. Then one computes ${\rm MMD}^2_u[\mathcal{F},X_i, Y_i]$ for each pair of blocks and takes an average. In online change-point detection, the sample size is not fixed. Our goal is to detect whether there is a change-point $\tau$ and its location, such that before the change-point, the samples are i.i.d. with a distribution $P$, and after the change-point, the samples are i.i.d. with a different distribution $Q$. New samples sequentially appear and we constantly test whether the incoming samples come from a different distribution. To reduce computational burden, in the online setting, we fix the block-size and adopt a sliding window approach. 

The detection statistic is constructed as follows. At each time $t$, we
treat the most recent $B_0$ samples as the post-change block, and there is a large amount of reference data. To utilize data efficiently, we utilize a common test block consisting of the most recent samples to form the statistic with $N$ different reference blocks. The reference blocks are formed by taking $NB_0$ samples without replacement from the reference pool. We compute ${\rm MMD}^2_u$ between each reference block with respect to the common post-change block, and take an average:
$$
Z_{B_0,t}=\frac{1}{N}\sum_{i=1}^{N}{\rm MMD}^2_u(X_i^{(B_0,t)},Y^{(B_0,t)}),
$$
where $B_0$ is the fixed block-size, $X_i^{(B_0,t)}$ is the $i$-th reference block at time $t$, and $Y^{(B_0,t)}$ is the the post-change block at time $t$. We divide each statistic by its standard deviation to form the online scan B-statistic:
$$
Z_{B_0,t}'=Z_{B_0,t}/({\rm Var}[Z_{B_0,t}])^{1/2}.
$$
The calculation of ${\rm Var}[Z_{B_0,t}]$ can be achieved using Theorem \ref{the:1}. The online change-point detection procedure is a stopping time: an alarm is raised whenever the detection statistic exceeds a pre-specified threshold $b>0$:
\begin{equation}
\label{eq:3.1}
T=\inf\{t:Z_{B_0,t}'>b\}.
\end{equation}

\begin{Th}
	\label{the:1}
	(Variance of $Z_B$ under the null). Given block size $B\ge2$ and the number of blocks $N$, under the null hypothesis,
	$$
		{\rm Var}[Z_B]=\tbinom{B}{2}^{-1}\left(\frac{1}{N}\mathbb{E}[h^2(x,x',y,y')]+\frac{N-1}{N}{\rm Cov}[h(x,x',y,y'),h(x'',x''',y,y')]\right),
	$$
	where $x,x',x'',x''',y$ and $y'$ are i.i.d. random variables with the null distribution $P$. 
\end{Th}

In the online setting, two commonly used performance metrics are \citep[see][]{xie2013sequential}: (1) the average run length (ARL), which is the expected time before incorrectly announcing a change of distribution when none has occurred; (2) the expected detection delay (EDD), which is the expected time to fire an alarm when a change occurs immediately at $\tau$ = 0. An accurate approximation to the ARL of online scan B-statistics is presented in Theorem \ref{the:2}. As a result, given a target ARL, one can determine the corresponding threshold value b from the analytical approximation, avoiding the more expensive numerical simulations. 

\begin{Th}
	\label{the:2}
	(ARL in online scan $B$-statistic). Let $B_0\ge2$. When $b\to\infty$, the ARL of the stopping time $T$ defined in \eqref{eq:3.1} is given by
	$$
		\mathbb{E}[T]=\frac{e^{b^2}/2}{b}\cdot\left\{\frac{2B_0-1}{\sqrt{2\pi}B_0(B_0-1)}\cdot v\left(b\sqrt{\frac{2(2B_0-1)}{B_0(B_0-1)}}\right)\right\}^{-1}\cdot\{1+o(1)\},
	$$
	where the special function 
	$$
	v(\mu)\approx\frac{(2/\mu)(\Phi(\mu/2)-0.5)}{(\mu/2)\Phi(\mu/2)+\phi(\mu/2)},
	$$
	$\phi(x)$ and $\Phi(x)$ are the are the probability density function and the cumulative distribution function of the standard normal distribution, respectively.
\end{Th}

\section{Set up}
\label{sec.exper}
We compare the scan B-statistic with two commonly used parametric statistics: the Hotelling’s $T^2$ and the generalized likelihood ratio (GLR) statistics. Assume samples $\{x_1,...,x_n\}$ and a fixed block-size $B_0=20$. At each time $t$, (1) we form a Hotelling’s $T^2$ statistic using the immediately past $B_0$ samples in $[t-B_0+1,t]$, $T^2(t)=B_0(\bar{x}_t-\hat{\mu}_0)^T\hat{\Sigma}_0^{-1}(\bar{x}_t-\hat{\mu}_0)$, where $\bar{x}_t=(\sum_{i=t-B_0+1}^{t}x_i)/B_0$, and $\hat{\mu}_0$ and $\hat{\Sigma}_0$ are estimated from reference data; (2) a GLR statistic can be constructed as $l(t)=t|\hat{\Sigma}_t|-(n-B_0)|\hat{\Sigma}_{t-B_0}|-B_0|\hat{\Sigma}_{t-B_0}^*|$, where $\hat{\Sigma}_{t-B_0}=(t-B_0)^{-1}(\sum_{i=1}^{t-B_0}(x_i-\bar{x}_i)(x_i-\bar{x}_i)^T)$ and $\hat{\Sigma}_{t-B_0}^*=B_0^{-1}(\sum_{i=t-B_0+1}^{t}(x_i-\bar{x}_i^*)(x_i-\bar{x}_i^*)^T)$. These two procedures detect a change-point whenever $T^2(t)$ or $l(t)$ exceeds a threshold for the first time. The thresholds for online scan B-statistic is obtained from Theorem \ref{the:2}, and from simulations for the Hotelling’s $T^2$ and GLR statistics.

To simulate EDD, let the change occur at the first point of the testing data. Consider the following cases:

\begin{itemize}
    \item \emph{Case 1} (mean shift): distribution shifts from $N(\mathbf{0}, I_{10})$ to $N(\mathbf{1}, I_{10})$;
    
    \item \emph{Case 2} (covariance change): distribution shifts from $N(\mathbf{0}, I_{10})$ to $N(\mathbf{0},\Sigma)$, where $[\Sigma]_{ii}=2, i=1, 2,...,5$ and $[\Sigma]_{ii} = 1, i = 6, \dots, 10$;

    \item \emph{Case 3} (covariance change): distribution shifts from $N(\mathbf{0}, I_{10})$ to $N(\mathbf{0}, 2I_{10})$;

    \item \emph{Case 4} (Gaussian to Gaussian mixture): distribution shifts from $N(\mathbf{0}, I_{10})$ to mixture Gaussian $0.3N(\mathbf{0}, I_{10})+0.7N(\mathbf{0}, 0.1I_{10})$;

    \item \emph{Case 5} (Gaussian to Laplace): distribution shifts from $N(0, 1)$ to Laplace distribution with zero mean and unit variance.
\end{itemize}

\section{Results}
\label{sec.result}

\begin{figure}[ht]
	\centering
	\includegraphics[width=\textwidth]{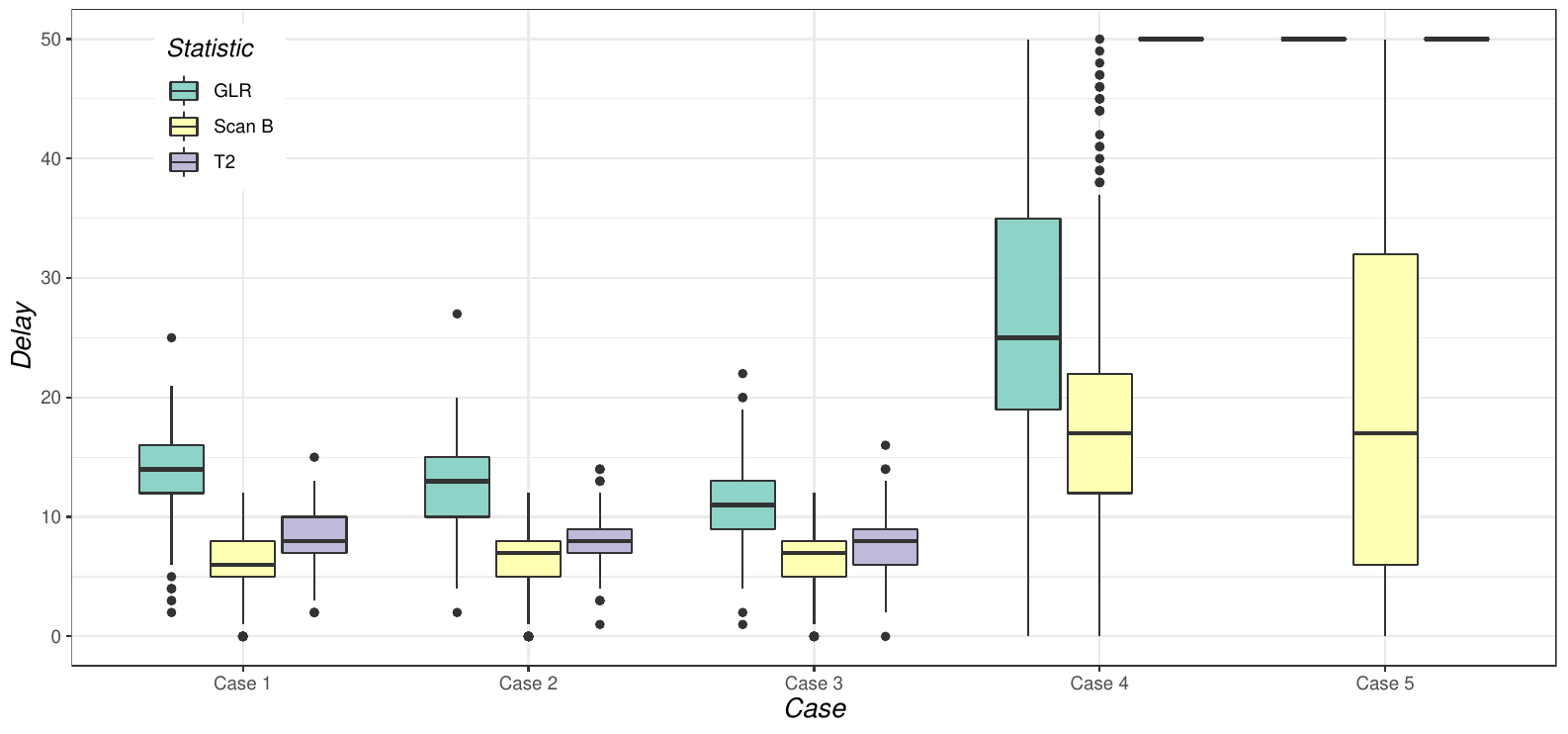}
	\caption{The detection delay of different statistics and different cases. The parameter is $B_0 = 20$ and thresholds for all methods are calibrated so that ARL$=5000$. The absence of boxplot means that the procedure fails to detect the change, i.e., EDD is longer than 50.}
	\label{fig5}
\end{figure}

We evaluate the EDD for each case using 500 Monte Carlo replications. The results are summarized in Figure \ref{fig5}. Note that in detecting changes in either Gaussian mean or covariance, the online scan B-statistic performs best compared with Hotelling’s $T^2$ and GLR, which is tailored to the Gaussian distribution. In the more challenging scenarios such as Case 4 and Case 5, the Hotelling’s $T^2$ and GLR may fail to detect the change-point whereas the online scan B-statistic can detect the change fairly quickly. 

Note that SBSK algorithm has many tuning parameters, such as the number of blocks $N$, the fixed block size $B_0$, and the kernel function $k(\cdot)$ and corresponding bandwidth $\sigma$. 

\begin{figure}[ht]
	\centering
	\includegraphics[width=\textwidth]{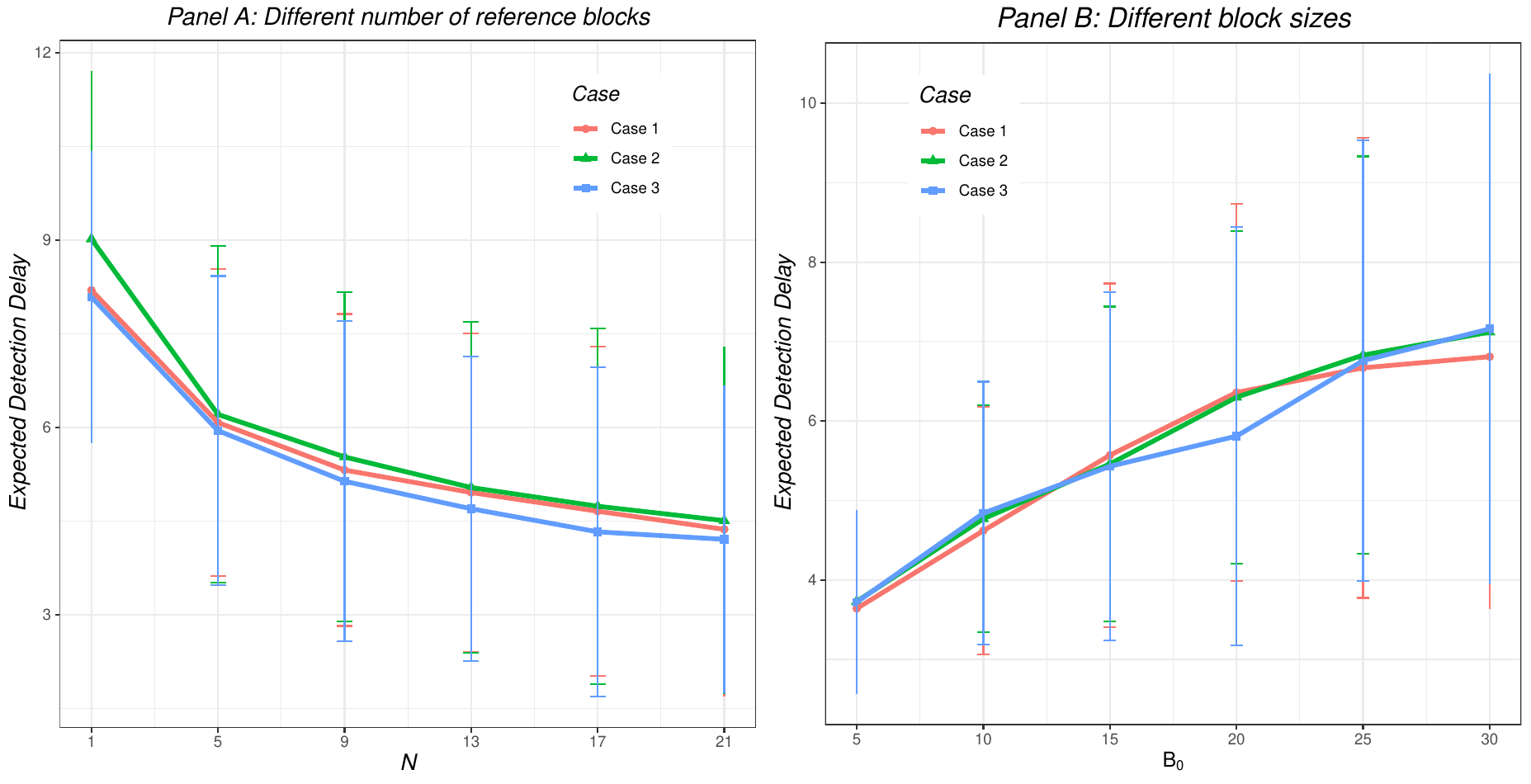}
	\caption{Comparison of EDD with different number of reference blocks $N$ (Panel A) and different block sizes $B_0$ (Panel B).}
	\label{fig6}
\end{figure}

In Figure \ref{fig6}, we show the comparison of EDD results with different $N$ (Panel A) and different $B_0$ (Panel B). Only Case 1, Case 2 and Case 3 introduced in Section~\ref{sec.exper} are used to compare. We can find that EDD decreases gradually with the increase of $N$. Intuitively, there are more reference blocks to compare with the post-change block, and the estimation of ${\rm MMD}^2_u$ is more accurate. When $B_0$ increases, the mean and standard deviation of EDD gradually increase. When we use a larger block size, we need more post-change samples from $Q$ to reflect the difference between the samples from $P$.

\begin{figure}[ht]
	\centering
	\includegraphics[width=\textwidth]{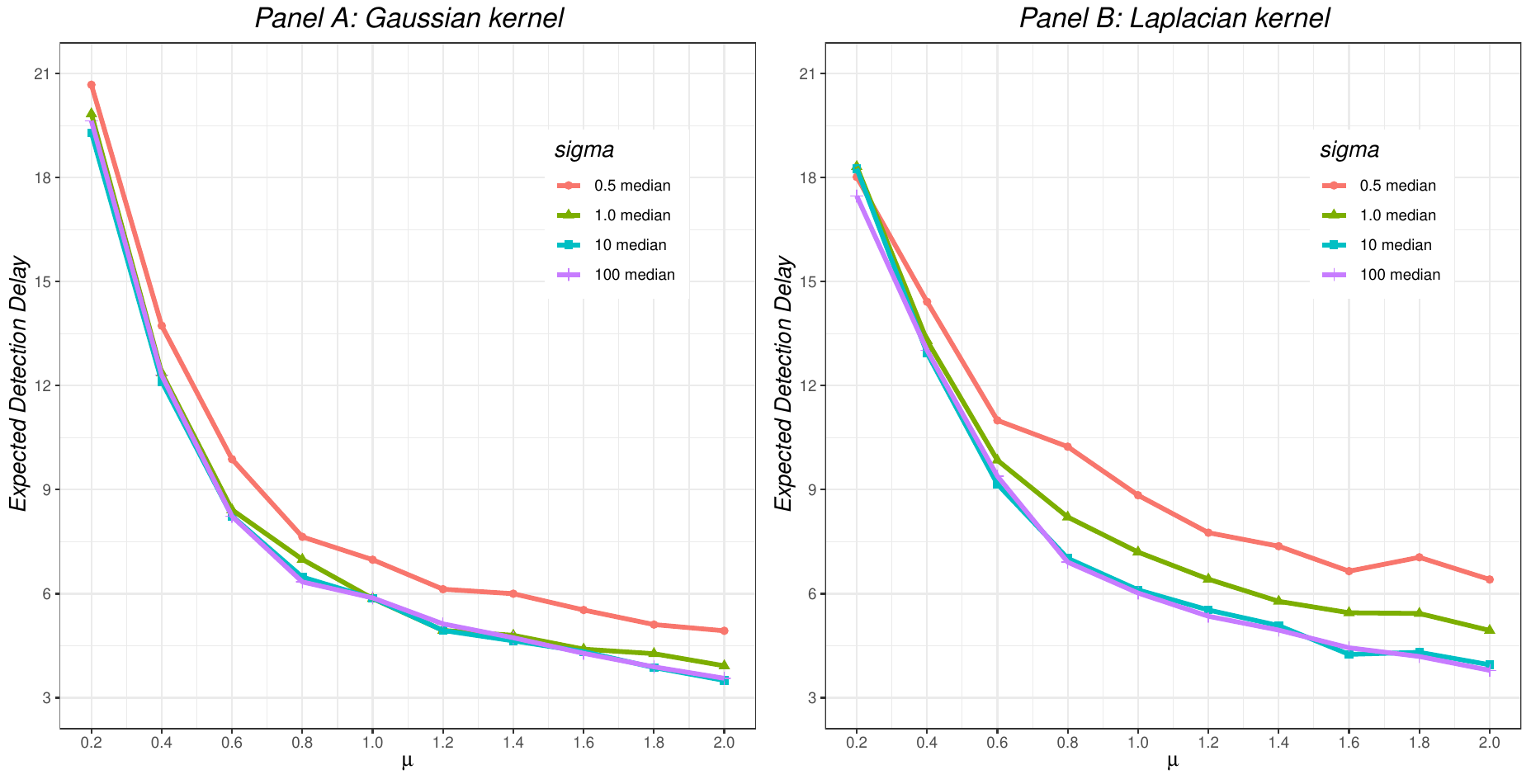}
	\caption{Comparison of EDD using Gaussian kernel (Panel A) and Laplacian kernel (Panel B) with different $\sigma$. Here, distribution shifts from $N(\mathbf{0}, I_{10})$ to $N(\mu\mathbf{1}, I_{10})$.}
	\label{fig7}
\end{figure}

In Figure \ref{fig7}, we illustrate the effect of different types of kernel and kernel bandwidth. For Gaussian RBF kernel and Laplacian RBF kernel, the bandwidth $\sigma > 0 $ is typically chosen using a “median trick” in \citet{scholkopf2002learning} and \citet{ramdas2015decreasing}, where $\sigma$ is set to be the median of the pairwise distances between data points. We find that Laplacian kernel performs slightly better than Gaussian kernel when $\mu$ is small, while the situation reverses when $\mu$ is large. The kernel bandwidth $\sigma=10\ {\rm median}$ and $100\ {\rm median}$ give relatively lower EDD.

\begin{figure}[ht]
	\centering
	\includegraphics[width=\textwidth]{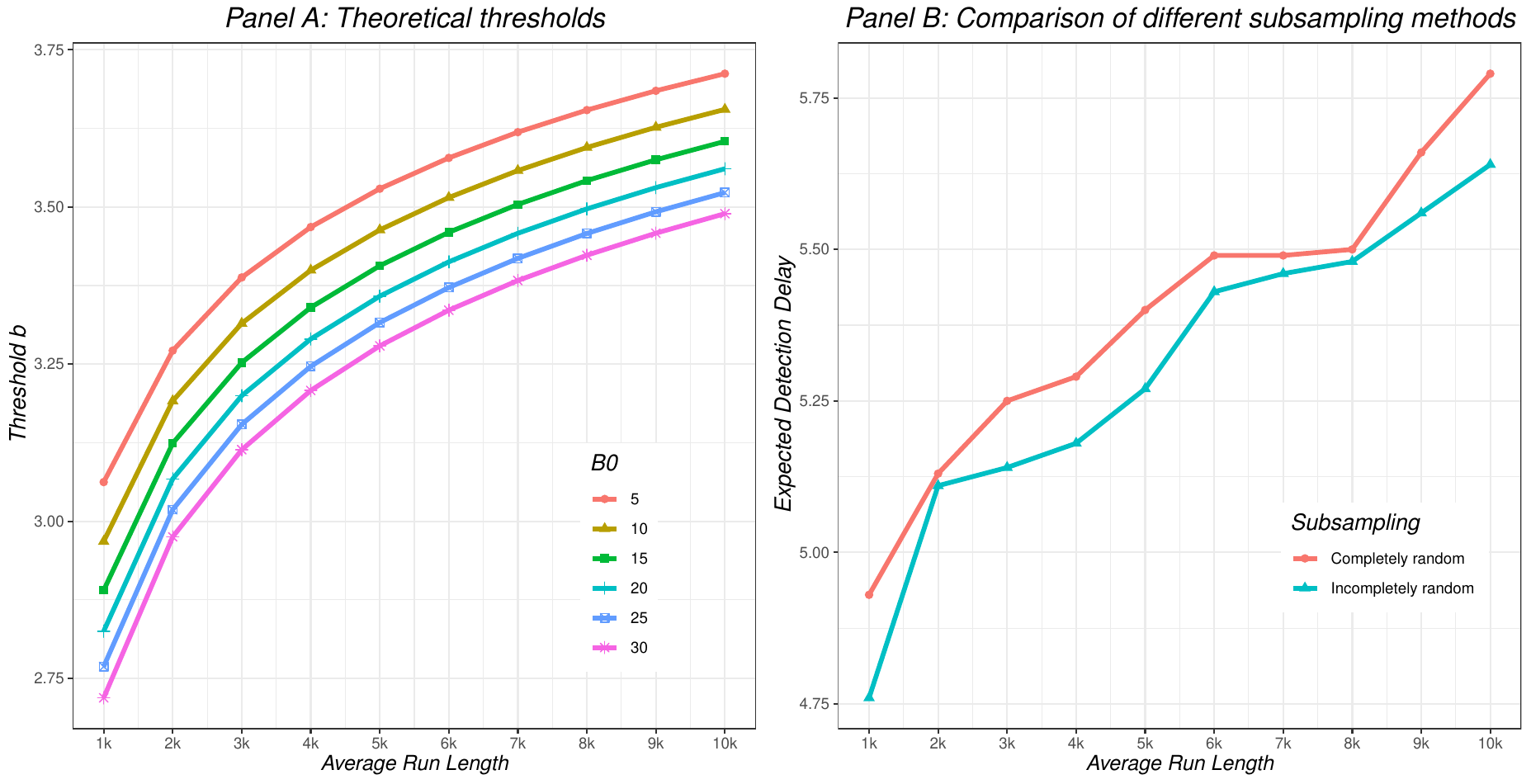}
	\caption{Comparison of completely random subsampling and using subsampling techniques when estimating the variance of $Z_B$. Given a range of target ARL values, thresholds determined from Theorem \ref{the:2} are shown in Panel A. Actually, $n=O\{(\log {\rm ARL})^{1/2}\}$. }
	\label{fig8}
\end{figure}

In the above discussion, a problem has been ignored. The variance of statistic $Z_B$ can be obtained according to Theorem \ref{the:1}, but we don't make any distribution assumption on $P$. The expectation and covariance can't be calculated theoretically, but can only be estimated by samples. However, if we try all the combination of 6 samples ($x,x',x'',x''',y,y'$) from 100 samples, there will be nearly $10^9$ choices. This computational complexity is unacceptable, so we can only do subsampling from the sample space. Completely random subsampling is not always the best choice, and we use subsampling techniques to make the selected samples uniformly  distributed. In Panel B of Figure \ref{fig8}, we show that EDD decreased after adding appropriate subsampling techniques. This problem is not discussed in the original paper, but our thinking in the process of algorithm reproduction, which may be a small improvement of the original SBSK algorithm.

\section{Conclusion}
\label{sec.con}

In this paper, the scan B-statistic is systematically introduced, and different number of blocks, block sizes, bandwidth and RBF kernels are compared in extensive numerical experiments. Result shows that the scan B-statistic achieves the best performance, and in the more challenging scenarios, parametric statistics may fail to detect whereas the scan B-statistic can detect the change fairly quickly. The use of subsampling techniques may be a small improvement of the original algorithm.

\clearpage
\bibliographystyle{apalike}
\bibliography{ref}

\end{document}